\documentclass[11pt,a4paper]{article}

\usepackage[utf8]{inputenc}
\usepackage[T1]{fontenc}
\usepackage{geometry}
\usepackage{authblk}  % Seamless handling of multiple authors and affiliations
\usepackage{hyperref} % For clickable links and emails

\usepackage{mathtools, amssymb, amsthm, bm}
\usepackage{graphicx}           % For including images
\usepackage{multirow}
\usepackage{hyperref}           % For hyperlinks
\usepackage{booktabs}           % For professional tables
\usepackage{caption}            % For customizing captions
\usepackage{subcaption}         % For subfigures
\usepackage{algorithm}          % For algorithms
\usepackage{algpseudocode}      % For algorithmic pseudocode
\usepackage{xcolor}             % For colored text
\usepackage{mathtools, amsmath, amssymb}          % Enhanced math capabilities
\usepackage{natbib}             % For references
\usepackage{pgffor} % Para iterar con \foreach
\usepackage{longtable} % Useful for large tables
\usepackage{array}
\usepackage[shortlabels]{enumitem}
\usepackage{newtxtext,newtxmath}

\usepackage{placeins}

\usepackage{pgffor}   % para \foreach
\usepackage{calc}     % para calcular anchos
\usepackage{rotating}
\usepackage{tikz}
\usetikzlibrary{positioning,fit,decorations.pathreplacing}

\setcitestyle{numbers}
\newcommand{\R}{\mathbb{R}}  % Real number symbol
\newcommand{\loss}{CoCo loss }
\newcommand{\Loss}{CoCo Loss }
\newcommand{\numClasses}{C}

\graphicspath{{../img/},{./img}}

\newtheorem{definition}{Definition}[section]

\title{\textbf{Contrastive-Collapsed Loss for Flexible and Geometrically Optimal Embeddings and Faster Convergence}}

\author[1]{Blanca Cano-Camarero\thanks{\texttt{blanca.cano@uam.es}}}
\author[1]{Ángela Fernández-Pascual\thanks{\texttt{a.fernandez@uam.es}}}
\author[1]{José R. Dorronsoro\thanks{\texttt{jose.dorronsoro@uam.es}}}

\affil[1]{Departamento de Ingeniería Informática, Universidad Autónoma de Madrid, Calle Francisco Tomás y Valiente, 11, 28049, Madrid, España}

\date{\today} % Or leave empty using \date{}

\begin{document}

\maketitle

% --- Abstract & Keywords ---
\begin{abstract}
In this work, we introduce CoCo, a loss function aimed at learning normalized and well-structured representations.
The proposed loss encourages intra-class collapse and inter-class contrast while preserving sufficient flexibility for neural networks to approximate geometrically optimal embeddings with large angular separation between classes.
We provide a theoretical analysis positioning CoCo with respect to related objectives such as dot regression and cross-entropy, showing that the new proposed loss benefits from closer initialization to the optimal configuration, more informative gradients, and stronger incentives for class-wise representation collapse.
Extensive experiments on diverse tabular datasets from the OpenML-CC18 benchmark show that CoCo achieves competitive performance with state-of-the-art methods, including kernel SVM, Random Forest, dot regression, and cross-entropy-based neural networks. 
In addition, both theoretical arguments and empirical analyses demonstrate that the proposal promotes tighter class clustering and faster convergence.
These results highlight CoCo loss as an effective objective for learning discriminative representations while maintaining competitive predictive performance.

\vspace{1.5em}
\noindent \textbf{Keywords:} Loss function, ETF, kernel, Neural collapse, Neural peeling, Early-stage convergence
\end{abstract}
\clearpage

% --- Main Text ---

\section{Introduction \label{sec:introduction}} 

The analysis of the penultimate layer of neural networks has been addressed by several 
prominent works 
In these studies, 
a ``collapse'' effect occurring in the last hidden layer during the terminal phase 
of training (TPT) has been observed and characterized. 
Initially, \cite{papyan2020prevalence} identified this phenomenon as \textit{Neural 
Collapse} (NC), noting that class means converge to the vertices of a Simplex 
Equiangular Tight Frame (ETF) while within-class variance diminishes. To provide 
a theoretical foundation, \cite{mixon2022neural} proposed the Unconstrained 
Features Model, which allows for a rigorous geometric analysis of the loss 
landscape. The robustness of these findings was further evaluated by 
\cite{andriopoulos2024prevalence}, who documented the prevalence of such 
structures across diverse training scenarios. 

Further synthesis of these principles was provided by \cite{kothapalli2022neural}, 
focusing on the relationship between feature collapse and the network's 
generalization capabilities. Recent extensions of this theory include the work 
of \cite{zhang2025all}, who analyze how these representations behave in high-dimensional 
scientific and multi-modal data, and \cite{rangamani2026deep}, who investigate the 
manifestation of collapse dynamics beyond the penultimate layer, reaching into 
the deeper architecture of the model.

The observed regularity of the penultimate layer has led several authors to 
propose a radical shift in architecture: dispensing with the trainable final 
classifier in favor of a fixed geometric structure 
\citep{zhu2021geometric}.
These works suggest that the network can directly learn 
embeddings that map to the vertices of a fixed \textit{Simplex Equiangular 
Tight Frame} (ETF).
The primary motivations for this approach are twofold. First, there are 
clear computational advantages, including a reduction in memory footprint 
and total parameter count \citep{hoffer2018fix}, alongside faster convergence 
to the optimal solution \citep{yang2022inducing}.
Second, the fixed 
structure acts as a strong inductive bias that encourages maximal class 
separation \citep{pernici2021regular}.
Furthermore, \cite{pmlr-v139-graf21a} demonstrate that this ETF structure is the optimal solution for supervised contrastive objectives, providing a theoretical bridge between contrastive loss and last-layer geometry.
However, these approaches 
constrain the embeddings to a fixed subspace, which may limit their flexibility and generalization.
Other approaches as \cite{ZHANG2025111400, ZHANG2022108877}, incorporate ETF information to modify the softmax loss for CNN problems.

Alternatively, other geometrical based methods already exist, such as kernel SVM (KSVM)~\citep{vapnik95, mavroforakis2006geometric, sanchez2003advanced}. 
The downside is that they rely on predefined kernels, typically Gaussian or polynomial, which may not adapt to the intrinsic nature of the data.
In fact, when the problem is complex and high-dimensional, such fixed kernels often fail to capture its underlying structure.
Moreover, these models usually depend on the kernel trick~\cite{scholkopf2000kernel} rather than an explicit lifting,
and who have a substantial computational computational cost.

To address the previously mentioned problems, this work introduces a new loss function, that we have named CoCo (\textit{Collapsing and Contrasting}) loss
whose minimization transforms the data to enforce normalization, intra-class attraction (collapsing) and inter-class separation (contrasting).
CoCo offers several key advantages:
\begin{description}
    \item[New properties for collapsing and contrasting data.] The data transformed by the solutions of our loss exhibit intra-class attraction, inter-class separation, and normalization.
    These effects induce the collapsing of data from the same class into a single vector, while contrasting different classes against each other.

    \item[Faster convergence.] We observe that CoCo models converge in fewer epochs than the other models.
       
    \item[Interpretable as a kernel.] The functions that minimize our loss act as liftings.
    Our approach can be viewed as a more flexible and potentially more expressive kernel, capable of producing richer feature representations.
\end{description}

This paper is structured as follows.
In Section~\ref{sec:introducing_coco} we introduce the proposed CoCo loss and then prove that it can be interpreted as a kernel. 
In Section~\ref{subsec:art_of_state} we position CoCo loss in the state of the art, including 
neural collapse, 
methods that remove the final classifier,
mathematical results on optimal geometric configurations and prediction accurateness.
In Section~\ref{sec:experiments} we conduct experiments to analyze the performance of the new proposed loss against other well-established methods.
Finally, in Section~\ref{sec:conclusions}
we provide some conclusions and further work.

\section{CoCo Loss
\label{sec:introducing_coco}}
\subsection{Definition}
In the context of classification, the primary objective of our proposed loss function 
is to learn a transformation $h$ that satisfies three key criteria: 
(1) it collapses intra-class samples into a single representative vector,
(2) maximizes inter-class separation, and 
(3) enforces a unit-norm constraint. 
This is achieved by requiring the transformed vectors to maintain an optimal angular distance, 
defined by their targets via a function $g$.
Consequently, $h$ maps the input data into 
a feature space that, by design, serve as the classification function itself.

\begin{definition}[Target similarity function]
\label{def:g}
Let $\numClasses \geq 2$ be the number of classes, $q$ denote the embedding dimension and $t$ the target dimension.  
We define the function $
g : \R^t \times \R^t \longrightarrow \R
$
as the target similarity between two labels.  
When $2 \leq \numClasses \leq q - 1$, $g$ is given by
\begin{equation}\label{eq:g_definition_etf}
g(y_i, y_j) =
\begin{cases}
\phantom{-\dfrac{1}{\numClasses }}\,1, & \text{if } y_i = y_j, \\
-\dfrac{1}{\numClasses - 1}, & \text{if } y_i \neq y_j .
\end{cases}
\end{equation}
This choice corresponds to the optimal angular geometry that maximizes inter-class separation.
\end{definition}

\begin{definition}[CoCo loss]
\label{def:coco}
In the context of supervised classification with $\numClasses$ classes, let
$X = \{x_i\}_{i=1}^{N} \subset \R^d$ be the input features,
$Y = \{y_i\}_{i=1}^{N} \subset \R^t$ the corresponding labels,
and let $h : \R^d \longrightarrow \R^q$ be the embedding function to be learned.
The \emph{CoCo loss} is defined as
\begin{equation}\label{eq:coco-loss}
\begin{aligned}
E_{\mathrm{coco}}(h)
= {} & 
\tfrac{2}{N (N+1)}\sum_{i=1}^{N}\sum_{j=i+1}^{N}
\left(
\langle h(x_i), h(x_j) \rangle - g(y_i, y_j)
\right)^2 \\
& + \tfrac{1}{N}\sum_{i=1}^{N}
\left(
\langle h(x_i), h(x_i) \rangle - g(y_i, y_i)
\right)^2 
\end{aligned}
\end{equation}
or equivalently
\begin{equation*}
\begin{aligned}
E_{\mathrm{coco}}(h)
= {} & 
\tfrac{2}{N (N+1)}\sum_{i=1}^{N}\sum_{j=i+1}^{N}
\left(
\langle h(x_i), h(x_j) \rangle - g(y_i, y_j)
\right)^2 
+ \tfrac{1}{N}\sum_{i=1}^{N}
\left(
\| h(x_i)\| - 1
\right)^2.
\end{aligned}
\end{equation*}
\end{definition}
To understand the motivation of our loss we first need the following definition: 
\begin{definition} \label{def:simplex_etf}
Let $\mathbf{S}$ be a $d \times \numClasses$ matrix whose columns are $\mathbf{s}_1, \dots, \mathbf{s}_{\numClasses}$.
The matrix $\mathbf{S}$ is called an ETF if it satisfies the following conditions:
\begin{enumerate}
    \item Each column has unit norm, i.e.
    $
    \|\mathbf{s}_n\|_2 = 1, \quad \text{for } n = 1, \dots, \numClasses.
    $
    \item The columns are equiangular, i.e., there exists a nonnegative constant $\alpha$ such that for all $m \neq n$ when $1 \leq m < n \leq \numClasses, 
    |\langle \mathbf{s}_m, \mathbf{s}_n \rangle| = \alpha.
    $
    \item The frame is tight, meaning that
    $
    \mathbf{S} \mathbf{S}^T = \frac{\numClasses}{d} P
    $, where $P\in \R^{d \times d}$ is a positive definite matrix.
    In the classical definition usually $P = \mathbf{I}_d$.
\end{enumerate}
If the entries of $\mathbf{S}$ are real numbers, we refer to $\mathbf{S}$ as a \textit{real ETF}.
Note that condition (3) implies $\numClasses \geq d$.
We will refer to it as a \textbf{$C$-simplex} when $C=d+1$.
\end{definition} 

ETFs arise as the solution to a classical geometric problem: the optimal arrangement of $\numClasses$ unit vectors in $\mathbb{R}^q$ that maximizes pairwise angular separation corresponds to the vertices of a regular $(\numClasses-1)$-dimensional simplex embedded in $\mathbb{R}^q$~\citep{el2004certain}.
In this configuration, all vectors are equiangular and achieve the maximum possible common angle.
For further details, we refer the reader to~\ref{appendix:CoCo_analysis}.
Within our framework, these vectors naturally play the role of class centers, toward which sample embeddings are encouraged to collapse.

\subsection{Properties of \Loss \label{sec:properties}}

Let $h: \R^d \longrightarrow \R^q$ be a mapping belonging to a hypothesis space $\mathcal{H}$.
If $h$ achieves a zero error in CoCo loss expression~\eqref{eq:coco-loss} on a given dataset $\mathcal{D} = \{(x_i, y_i)\}_{i=1}^N$, several relevant observations can be drawn:
\begin{itemize}
    \item \textbf{Normalization}: For each $i$, if $j = i$, then  $g(y_i,y_j) = 1$, so
    \begin{equation*}
        \left(\langle h(x_i ), h(x_i) \rangle -g(y_i,y_i)\right)^2 
        = 
        (\|h(x_i )\|^2 -1)^2
    \end{equation*}
    and therefore,  if a zero minimum of $E_{coco}$ is reached, then $\|h(x_i )\|  = 1$ for each sample.

    \item \textbf{Intra-class attraction}. 
When two samples belong to the same class, i.e., $y_i = y_j$, we have $g(y_i, y_j) = 1$ (see Definition~\ref{def:g}).
Consequently, since $\|h(x_i)\| = \|h(x_j)\| = 1$, it follows that $\langle h(x_i), h(x_j) \rangle = 1$, implying that the angle between them is zero.
Thus, minimizing CoCo loss forces their embeddings to coincide, yielding $h(x_i) = h(x_j)$.
This property causes the representations within each class to collapse into a single point on the unit hypersphere.

    \item \textbf{Inter-class separation.}
Let $i$ and $j$ be indices of samples belonging to different classes.
As given in Definition~\ref{def:g}, $g(y_i, y_j) = \gamma^\star = \frac{-1}{\numClasses - 1}$, where $\gamma^\star$ represents the scalar product corresponding to the maximum equiangular configuration admissible in $\mathbb{R}^q$ for $\numClasses$ unit-norm vectors.
Since $\|h(x_i)\| = \|h(x_j)\| = 1$, it follows that $\cos(\measuredangle(h(x_i), h(x_j))) = \gamma^\star$.
As all samples within each class collapse to a single point, different classes are therefore separated by an angle of $\arccos(\gamma^\star)$, generating a well-defined geometric contrast between them.
\end{itemize}

\begin{figure}[t]
    \centering
    \begin{subfigure}{0.24\textwidth}
        \includegraphics[width=\linewidth]
        {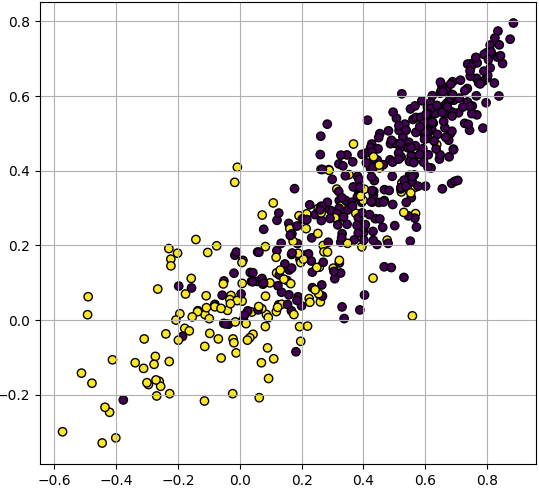}
        \caption{50 epochs}
    \end{subfigure}%
    \hfill
    \begin{subfigure}{0.24\textwidth}
        \includegraphics[width=\linewidth]{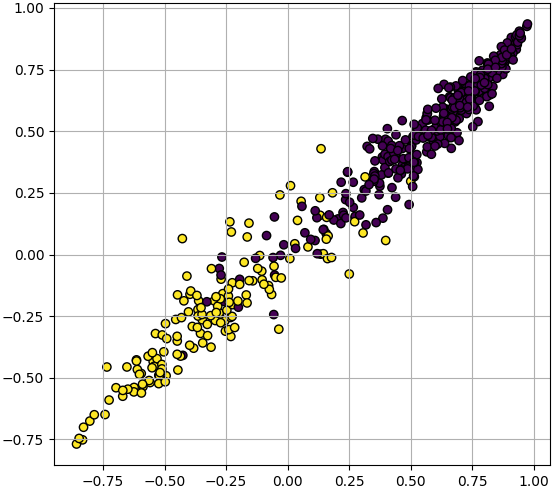}
        \caption{150 epochs}
    \end{subfigure}%
    \hfill
    \begin{subfigure}{0.24\textwidth}
        \includegraphics[width=\linewidth]{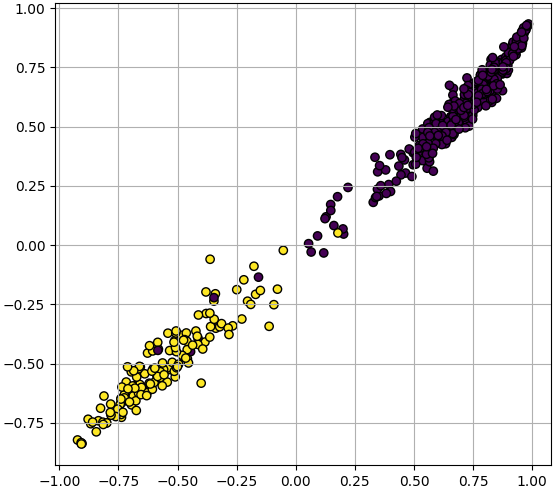}
        \caption{250 epochs}
    \end{subfigure}%
    \hfill
    \begin{subfigure}{0.24\textwidth}
        \includegraphics[width=\linewidth]{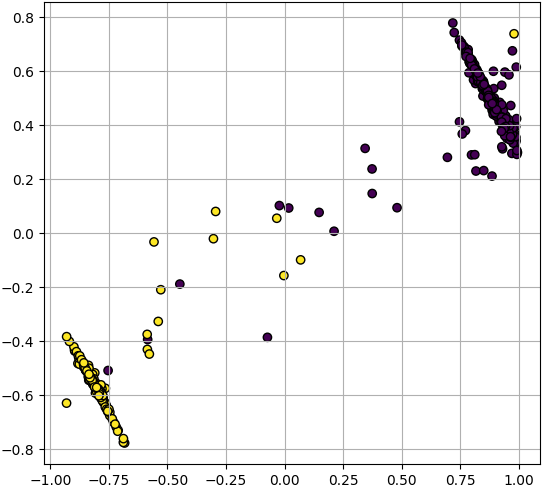}
        \caption{500 epochs}
    \end{subfigure}
    \caption{
        Evolution of the predictions for a neural network with a 2-dimensional output, as the number of training epochs increases, trained using CoCo loss function on German Credit Data with $\alpha = 1$. 
    \label{fig:evolution-stalog}
    }
\end{figure}

As an example of this effect, we trained a neural network with one hidden layer of 1024 neurons, using the loss function~\eqref{eq:coco-loss} and an output dimensionality of $2$ on the UCI German Credit dataset \citep{statlog_(german_credit_data)_144}. The results, shown in Figure~\ref{fig:evolution-stalog}, illustrate the progression of class separation over the training epochs.
Blue points represent one class, while yellow points represent the other.
As training progresses and the \loss error decreases, the two classes become increasingly separated, while the intra-class data points become more tightly clustered.
It can also be visually observed in the plot that the yellow class concentrates around the vector $(-0.75,-0.60)$, whereas the blue class concentrates around $(0.50,0.85)$. The corresponding norms are $|(-0.75,-0.60)|\approx 0.960$ and $|(0.50,0.85)|\approx 0.986$. The angle between these two vectors is
$$\theta = \arccos\left(\frac{\langle (-0.75,-0.60),(0.50,0.85)\rangle}{\|(-0.75,-0.60)\|\,\|(0.50,0.85)\|}\right)
\approx 2.777~\text{rad} \approx 0.884\,\pi,$$
thus the cosine is
$\cos\theta \approx -0.934$,
which is close to the theoretical target $\gamma^\star = -1$.

However, due to the simplicity of the neural network architecture and the fact that the data are being projected onto a two-dimensional space, the classes do not collapse perfectly in practice.

Some interesting observation about CoCo can be made:
\label{subseb:obsevation_coco}
\begin{enumerate}
  
    \item \textbf{CoCo loss liftings $h$ induce a kernel.} 
    The function $k(x_i, x_j) := \langle h(x_i), h(x_j)\rangle$ is by definition a kernel. 
    This approach is particularly interesting because it offers a way to design kernels that reflect the intrinsic structure of the data, providing an alternative to the conventional methods~\cite{LOPEZSANCHEZ2018130} that forces the user to fix a kernel~\cite{FANG2023109057} and evaluate its performance instead of building it directly from the data. 

    \item \textbf{If $q = \numClasses$, is CoCo loss equivalent to cross entropy?} The short answer is no, but it is closely related.
    Compared to state-of-the-art models, CoCo does not have a geometric structure that replaces labels (see \cite{hoffer2018fix, pernici2021regular, yang2022inducing}).
    Instead, based on the intrinsic structure of the data, it converges to non-fixed vectors that satisfy the properties described in Section \ref{sec:properties}.
    However, as CoCo enforces intra-class attraction and inter-class separation, if a representative of each class is known, computing predictions becomes straightforward.
    
    \item \textbf{Computational cost.}
    Let $n$ denote the batch size. If $h$ is implemented as a neural network, CoCo loss involves $\tfrac{n(n-1)}{2}$ pairwise terms, whereas the cross-entropy loss contains only $n$.
    Therefore, for the same batch size, CoCo loss exhibits a quadratic computational cost with respect to $n$, in contrast to the linear cost of cross-entropy.

    \item \textbf{Extreme case for output dimension $q$ and the number of classes $\numClasses$.}
    What happens when $q <  \numClasses -1$ and $\numClasses > 2$?
    First, this is a special case in which $h$, minimum of \eqref{eq:coco-loss}, rather than a lifting works as a projection.
    Second, it does not always admit a real (or even complex) solution to define a ETF (Definition \ref{def:simplex_etf}) as \cite{sustik2007existence} proved.
    However, if a solution exists, the optimal scalar product is given by \cite{welch2003lower}: 
\begin{equation}\label{eq:targets-relation-welch-bound}
  g(y_i, y_j) = \left\{
    \begin{array}{ccc}
        1 &if& y_i = y_j, \\
       \sqrt{ \frac{\numClasses-q}{q(\numClasses-1)}} & if & y_i \neq y_j.
    \end{array}
    \right.
\end{equation}

    \item \textbf{Can $g$ be defined without explicitly depending on the number of classes? }
Yes. Assuming the output dimension $q$ of $h$ is large enough to have a real solution 
(i.e. implicitly satisfies $q \geq \numClasses - 1$),
observe that for two numbers of classes $C_1 > C_2$, the target similarity function defined in Definition \ref{def:g} for $C_1$, $g_{c_1}$, is also a valid choice for the $C_2$ problem: it still satisfies the collapsing, contrasting, and normalization properties, although the resulting angle is not maximal. Taking this idea further, letting $\numClasses \rightarrow \infty$ in \eqref{eq:g_definition_etf} yields a fully class-independent limit $\hat g$ defined as
\begin{equation}\label{eq:targets-relation-orthoplex}
  \hat g(y_i, y_j) = \left\{
    \begin{array}{ccc}
        1 & if & y_i = y_j ,\\
        0 & if & y_i \neq y_j.
    \end{array}
    \right.
\end{equation}

\end{enumerate}

\subsection{Balanced \Loss}
In some situations, when the dataset classes are highly imbalanced, it would be desirable to adjust the loss in order to obtain better results~\cite{gao2026comprehensive, zhang2025all}. 
Even though an under-sampling or over-sampling can be used, it is often easier to adapt the loss in the following way:
\begin{equation}
    \begin{split}
        \label{eq:balanced-coco-loss} 
        E_{bal\,CoCo}(h) &= %\alpha 
        \sum_{i = 1}^{N} \sqrt\frac{1}{\pi_i}\sum_{j = i+1}^{N}
        \left[
            \sqrt\frac{1}{\pi_j}
            \left(
                \langle h(x_i),h(x_j) \rangle - g\left(y_i, y_j\right)
            \right)^2 
        \right]\\
        &+  \sum_{i = 1}^{N} \frac{1}{\pi_i}
        \left(
            \langle h(x_i),h(x_i) \rangle - g\left(y_i, y_i\right)
        \right)^2,
    \end{split}  
\end{equation}
where $\pi_i$ is the prior of class $y_i$.

This modification can be interpreted as an approximation of the expected loss under a reweighted experimental prior distribution of the labels.
Hence, instead of computing a uniform average over all pairs, the loss performs a weighted expectation where minority classes contribute more significantly, compensating for class imbalance in the dataset.

\subsection{Classification using \Loss}
\label{sub:prediction_with_coco}
A solution $h: \R^d \longrightarrow \R^q$ of CoCo loss (as defined in \eqref{eq:coco-loss} or \eqref{eq:balanced-coco-loss}) can be interpreted as a feature transformation function that do not depend on any prefixed kernel.
Consequently, it follows a similar downstream usage pattern to common feature mapping methods such as Nyström~\citep{williams2000using} or Random Fourier Features~\citep{rahimi2007random}. 

The training and inference procedure is structured as follows:
\begin{enumerate}
    \item \textbf{Training the Transformation:} Define and train a neural network $h$ using one of the CoCo variants, \eqref{eq:coco-loss} or \eqref{eq:balanced-coco-loss}, as the objective function. This process optimizes the network to map inputs into the ETF-structured feature space.
    
    \item \textbf{Supervised Classification:} Once $h$ is trained, the dataset is transformed into $h(X)$. Since $h$ induces a potentially linearly separable space by design, classification can be performed using one of the following strategies:
    \begin{enumerate}
        \item \textbf{Linear Classifiers:} Given the geometric properties of the transformed space, a simple linear model, such as Ridge Regression or a linear SVM, is sufficient to achieve high discriminative performance.
        
        \item \textbf{Centroid-based Inference:} Alternatively, one can compute class means in the transformed feature space as:
        \[
        \mu_c = \frac{1}{N_c} \sum_{i: y_i = c} h(x_i),
        \]
        where $N_c$ is the number of samples belonging to class $c$. For a query vector $x$, the predicted label is assigned to the class $c$ whose centroid $\mu_c$ is closest to $h(x)$, typically measured by Euclidean or angular distance.
        \item \textbf{Relying on a Gaussian Mixture Model (GMM):}  We assume that the feature representations $h(x)$ of samples belonging to class $c$ follow a multivariate normal distribution centered at the class centroid $\mu_c$ with covariance $\Sigma_c$.

During the training phase, we estimate these parameters via Maximum Likelihood Estimation (MLE). Since the class labels are known, the estimation is decoupled, yielding the centroid (mean) $\hat{\mu}_c$, the covariance matrix $\hat{\Sigma}_c$, and the class prior $\hat{\pi}_c$ as follows:
\begin{equation}
    \hat{\mu}_c = \frac{1}{N_c} \sum_{i = 1 }^{N_c} h(x_{i,c}), \quad 
    \hat{\Sigma}_c = \frac{1}{N_c} \sum_{i = 1 }^{N_c}  (h(x_{i,c}) - \hat{\mu}_c)(h(x_{i,c}) - \hat{\mu}_c)^T, \quad 
    \hat{\pi}_c = \frac{N_c}{N}
\end{equation}
if $\Lambda_c = \{x_i \in X : y_i = c\}$ then 
 $x_{i,c}$ represents the $i-$element of $\Lambda_c$ and  $N_i = |\Lambda_c|$, and $N$ is the total number of training samples.

For the prediction of a new sample $x$, we compute the posterior probability of it belonging to class $c$ using Bayes' rule:
\begin{equation}
    p(c \mid h(x)) = \frac{\hat{\pi}_c \,\mathcal{N}(h(x) \mid \hat{\mu}_c, \hat{\Sigma}_c)}{\sum_{j=1}^{C} \hat{\pi}_j \mathcal{N}(h(x) \mid \hat{\mu}_j, \hat{\Sigma}_j)}.
\end{equation}
Finally, the predicted label corresponds to the class maximizing this posterior probability:
$\hat{y} = \operatorname*{argmax}_{c \in \{1, \dots, C\}} p(c \mid h(x))$

    \end{enumerate}
\end{enumerate}

By design, the transformation $h$ acts as a \textit{classifier function} that reconfigures the data manifold. This ensures that even the simplest linear decision boundaries can effectively partition the classes in the $q$-dimensional embedding space.

\section{Positioning CoCo Loss in the State of the Art}
\label{subsec:art_of_state}
Supervised classification has traditionally been governed by two paradigms: the \textit{probabilistic} approach, typified by Cross-Entropy (CE)\citep{hastie2009elements, mao2023cross}, and the \textit{geometric} approach, exemplified by SVMs.
While CE focuses on estimating posterior distributions, KSVMs rely on the kernel trick to map data into a high-dimensional Reproducing Kernel Hilbert Space (RKHS) where linear separation is possible.
However, KSVMs often rely on fixed, predefined kernels, leading to rigid geometries that do not adapt to the data's intrinsic structure.

Recent studies on \textit{Neural Collapse} (NC) \citep{papyan2020prevalence,hong2024neural} have bridged these views, experimentally demonstrating that at the terminal phase of training (near-zero training error), CE-based models converge to a highly structured geometric configuration, more specifically, an ETF.
This suggests that optimal classification is inherently geometric.
While recent works such as Dot-Regression (DR) \citep{yang2022inducing} or fixed-pattern models \citep{wang2024navigate, hoffer2018fix, pernici2021regular} attempt to target this geometry directly to accelerate convergence, they typically enforce alignment to fixed, predefined embeddings.

In this landscape, CoCo loss emerges as a data-adaptive geometric framework.
Unlike DR, CoCo does not impose a static target; instead, it identifies and adapts the optimal geometric configuration for the specific problem at hand. Furthermore, while KSVMs fix the Hilbert space \textit{a priori}, CoCo effectively learns a data-driven kernel. The resulting mapping $h$ induces a lifting into a finite-dimensional space where the problem becomes, by definition, potentially linearly separable.
Thus, CoCo serves as both, a feature transformation and the classification function itself, combining the interpretability of geometric margin-based methods with the flexibility of modern deep learning architectures.

Consequently, our benchmarks will compare CoCo against CE for probabilistic baselines, KSVM for fixed-kernel geometric baselines, and DR for fixed-embedding collapse models.

\subsection{CoCo and Neural Collapse}
Let $h_i \in \mathbb{R}^d$ be the feature representation of a sample $x_i$, and $w_c$ the classifier weight vector for class $c$.
NC is characterized by:
\begin{description}
    \item[NC1] \textit{Within-Class Variability Collapse}: Features belonging to the same class concentrate around their class mean, i.e., $h_i \approx \mu_c \; \forall i$ s.t. $y_i = c$.
    \item[NC2] \textit{Convergence to Simplex ETF}: The class means $\{\mu_c\}_{c=1}^{\numClasses}$ become equiangular and have equal norms, forming a simplex ETF where 
    $
    \frac{\mu_c^\top \mu_j}{\|\mu_c\|\|\mu_j\|} = -\frac{1}{\numClasses-1}, \; \forall c \neq j.
    $
    \item[NC3] \textit{Self-Duality}: The classifier weights align with the class means: $\frac{w_c}{\|w_c\|} = \frac{\mu_c}{\|\mu_c\|}$.
    \item[NC4] \textit{Nearest Class Mean}: The classifier simplifies to a nearest-mean decision rule.
\end{description}

It can be shown that any solution $h$ that minimizes CoCo loss (with the loss closes to zero), satisfies the NC phenomenon.

In particular, taking into account the properties in Section~\ref{sec:properties}:
\begin{description}
    \item[NC1 \& NC4] For any two samples $x, x'$ of the same class $c$, CoCo loss enforces $\langle h(x), h(x') \rangle = 1$. Given the unit-norm constraint, this implies $h(x) = h(x')$. Consequently, all samples collapse to the class mean ($h(x) = \mu_c$), satisfying NC1 and therefore leading to the nearest-class-mean rule NC4.
    \item [NC2] emerges naturally from the definition of the function $g$ and the optimal constant $\gamma^*$.
    By setting the inter-class target to $\gamma^* = -1/(\numClasses-1)$, the transformation $h$ is explicitly optimized to produce a simplex ETF structure.
    \item [NC3] While CoCo does not explicitly define a final linear model, the optimal classifier that maximizes the separation margin in the feature space is given by $w_c \propto \mu_c$.
    Since all embeddings $h(x)$ are constrained to the unit hypersphere and the classification is performed in the angular domain, the decision boundaries naturally align with the class prototypes, thereby satisfying the self-duality/alignment property.
\end{description}

\subsection{Positioning CoCo with Respect to DR \label{sub:positioning_coco}} 
Finally, note that the key difference between CoCo and DR is that DR directly enforces NC properties by aligning features with fixed target vectors $\{v_c\}$ through the loss $\mathcal{L}_{\mathrm{DR}} = \frac{1}{2 N} \sum (h_i \cdot v_{y_i} - 1)^2$.
While both DR and CoCo aim to obtain a simplex ETF structure, they differ in the following aspects:
\begin{itemize}
    \item \textbf{Discovery vs. Imposition:} DR fixes targets $v_c$ beforehand; CoCo discovers them through pairwise relations $\langle h(x_i), h(x_j) \rangle = g(y_i, y_j)$.
    \item \textbf{Complexity:} DR uses $\mathcal{O}(N)$ constraints (sample-to-prototype), whereas CoCo uses $\mathcal{O}(N^2)$ (full pairwise consistency).
    \item \textbf{Adaptability:} DR requires balanced classes and fixed targets; CoCo adapts its geometric configuration to the inherent structure of the data.
\end{itemize}

It is worth noting that CoCo presents two main advantages over DR. 
First, CoCo may converge faster than DR. This effect arises from two complementary factors: a closer initialization to the optimal solution (see Section~\ref{subsub:closer_initialization}) and more informative gradients due to the local curvature of the loss landscape (see Section~\ref{subsub:local_curvature}). 

Second, CoCo exerts a stronger effect in both collapsing representations within the same class and separating representations across different classes. 
As shown in Section~\ref{subsub:class_dispersion}, this property not only distinguishes CoCo from DR but also provides additional insight into the underlying geometry induced by the CoCo objective.

\subsubsection{Closer initialization to the optimal solution\label{subsub:closer_initialization}}
Let $h:\mathbb{R}^d \rightarrow \mathbb{R}^q$ be a neural network randomly initialized. 
For a sample $x$, the output representation $h(x)$ can be regarded as a random vector following some distribution $\mathcal{Q}$.

In the case of DR, once the ETF vectors $v_1,\dots,v_C \in \mathbb{R}^q$ are fixed, the optimal solution for a sample $x_i$ of class $c$ is characterized by $(h(x_i)^\top \cdot v_c - 1)^2 = 0 .$
Therefore the set of optimal representations for class $c$ is
$H_{DR}^c = \{ h \in \mathbb{R}^q : h^\top v_c = 1 \}.$
This set is an affine hyperplane of dimension $q-1$.
The distance from a point $z$ to this hyperplane is
$\mathrm{dist}(z,H_{DR}^c)
=
\frac{|v_c^\top z - 1|}{\|v_c\|}.
$
Since ETF vectors are normalized, i.e. $\|v_c\|=1$, the expected distance from a random initialization is

\begin{equation}\label{eq:dist_dr}
\mathbb{E}[\mathrm{dist}(h,H_{DR})]
=
\lambda_q(\mathcal{Q})^{-1}
\int_{\mathcal{Q}} |v_c^\top h(x) -1| \, dh ,
\end{equation}
where $\lambda_q$ denotes the Lebesgue measure.

In contrast, CoCo does not impose any predefined direction for the representations. 
The only constraint is unit normalization of the embedding, so the solution set is the unit hypersphere 
$H_{\text{CoCo}}
=
\{ h \in \mathbb{R}^q : \|h(x)\|_2 = 1 \}.
$
The distance from a point $z$ to this set is
$\mathrm{dist}(z,H_{\text{CoCo}})
=
|\|z\|_2 - 1|.
$
Thus the expected distance is
\begin{equation}\label{eq:dist_coco}
\mathbb{E}[\mathrm{dist}(h,H_{\text{CoCo}})]
=
\lambda_q(\mathcal{Q})^{-1}
\int_{\mathcal{Q}} |\|h(x)\|_2 - 1| \, dh .
\end{equation}
Using Cauchy--Schwarz and the fact that $\|v_c\|=1$ we obtain
$
|v_c^\top h(x)|
\le
\|h(x)\|_2 .
$
Hence $|v_i^\top h(x)-1|
\ge
|\|h(x)\|_2-1|
-
|v_i^\top h(x) - \|h(x)\|_2|$
and it follows that
$$|v_c^\top h(x) - 1|
\ge
|\|h(x)\|_2 -1|.$$
Comparing~\eqref{eq:dist_dr} and~\eqref{eq:dist_coco} we obtain
\[
\mathbb{E}[\mathrm{dist}(h,H_{\text{CoCo}})]
\le
\mathbb{E}[\mathrm{dist}(h,H_{DR})].
\]
This shows that, under random initialization, the representations are on average closer to the CoCo solution set than to the DR solution set.
The previous argument is derived for a single sample.
When multiple samples of the same class are considered, compatibility between their representations must be taken into account.
We assume a mild smoothness condition on the embedding function, ensuring that samples from the same class produce nearby representations.
Under this assumption, samples that are close in the input space produce nearby embeddings.
Since samples from the same class typically lie in a compact region of the input space, their representations remain close during training.
Consequently, the pairwise constraints introduced by CoCo do not introduce incompatibilities and instead promote a consistent clustering of same-class embeddings.

\subsubsection{Local curvature of the loss\label{subsub:local_curvature}}

To better understand the convergence behavior of DR and CoCo, we analyze the local geometry of their loss functions around the optimal solution. A standard way to characterize this geometry is through the Hessian matrix of the loss with respect to the representations.
The Hessian captures the local curvature of the optimization landscape and determines how the loss changes in different directions around the optimum.
The eigenvalues of the Hessian describe the curvature along the corresponding eigen-directions. Large eigenvalues indicate directions with strong curvature, where the loss changes rapidly, while small eigenvalues correspond to flat directions where the loss varies slowly. If some eigenvalues are zero, the loss surface contains completely flat directions and the solution is not isolated.

In particular, the condition number of the Hessian summarizes the geometry of the landscape. For a normal positive semi-definite matrix $H$, it is defined as
\begin{equation}
\kappa(H) = \frac{\lambda_{\max}(H)}{\lambda_{\min}(H)},
\end{equation}
where $\lambda_{\max}$ and $\lambda_{\min}$ denote the largest and smallest non-zero eigenvalues of $H$. The condition number measures how anisotropic the curvature of the loss landscape is. When $\kappa(H)$ is close to $1$, the curvature is similar in all directions and the landscape resembles a well-shaped quadratic bowl. In contrast, a large condition number indicates highly elongated valleys, where optimization methods based on gradient descent tend to progress slowly.
Therefore, analyzing the spectrum of the Hessian provides insight into the difficulty of the optimization problem. Well-conditioned problems lead to stable and fast convergence, whereas ill-conditioned or degenerate landscapes can significantly slow down training. 
In the following, we compare the Hessian structures induced by DR and CoCo and show that they lead to fundamentally different conditioning properties of the optimization landscape.

A useful way to compare the convergence behaviour of DR and CoCo is to study the local curvature of their loss functions near the optimal configuration. This can be analyzed through the Hessian with respect to the representations.

%\paragraph{Dot Regression}
For a sample $x_i$ belonging to class $c$, the DR loss is
$L_{DR} = (h_i^\top v_c - 1)^2 .$
The gradient with respect to the representation is $\nabla_{h_i} L_{DR} = 2(h_i^\top v_c - 1)v_c$. 
Differentiating again yields the Hessian
$\nabla_{h_i}^2 L_{DR} = 2 v_c v_c^\top .$
This matrix has rank $1$. Its eigenvalues satisfy
\begin{equation}
\lambda_1 = 2, \qquad \lambda_2=\dots=\lambda_q = 0 .
\end{equation}
Therefore the loss exhibits curvature only along the direction $v_c$, while the $q-1$ orthogonal directions are completely flat.
These flat directions correspond to the affine solution set
$H_{DR}^{v_c}$.
As a consequence, the Hessian is degenerate and the condition number satisfies $\kappa(H_{DR}) = \infty$, indicating a highly ill-conditioned optimization landscape.

For two samples $i,j$, the CoCo loss takes the form
$L_{CoCo} = (h_i^\top h_j - g_{i j})^2.$

The gradient with respect to $h_i$ is $\nabla_{h_i} L_{CoCo} = 2(h_i^\top h_j - g_{i j})h_j$, and differentiating again gives
$\nabla_{h_i}^2 L_{CoCo} = 2 h_j h_j^\top.$
When summing the contributions of all pairs in the loss, the resulting Hessian becomes
$H_{CoCo} = 2\sum_j h_j h_j^\top .$
To better interpret the structure of the Hessian, observe that
the matrix $H_{CoCo}$ can be related to the second moment matrix of the embeddings. 
Indeed,$\frac{1}{N}\sum_{j} h_j h_j^\top = \mathbb{E}[hh^\top],$
where $\mathbb{E}[hh^\top]$ denotes the second moment of the representation distribution. 
If the embeddings are centered, this quantity coincides with their covariance matrix.

Near the optimal configuration, the representations follow the ETF geometry observed in Neural Collapse: embeddings have unit norm and are symmetrically distributed across directions of $\mathbb{R}^q$. 
Under this symmetry assumption, no direction in the space is preferred, which implies that the second moment matrix must be proportional to the identity, $\mathbb{E}[hh^\top] = \alpha' I_q.$

The constant $\alpha'$ can be determined using the trace. Since $\|h\|=1$,
$
\mathrm{trace}(hh^\top) = \|h\|^2 = 1 .
$
Taking expectation gives
$\mathrm{trace}(\mathbb{E}[hh^\top])$. 
Substituting $\mathbb{E}[hh^\top] = \alpha' I_q$ yields
$q\alpha' = 1,$
which implies $\mathbb{E}[hh^\top] = \frac{1}{q} I_q .$
Therefore,
$
\sum_{j} h_j h_j^\top \approx \frac{N}{q} I_q ,
$
and the Hessian becomes approximately
$H_{CoCo} \approx 2\frac{N}{q} I_q .$

This shows that the curvature of the CoCo loss is approximately isotropic near the optimum, meaning that all directions of the representation space have comparable curvature. This Hessian has full rank and eigenvalues of comparable magnitude, yielding a bounded condition number $\kappa(H_{CoCo}) = O(1)$.

These observations reveal a fundamental difference between the two losses. DR exhibits $q-1$ flat directions near the optimum, producing an ill-conditioned optimization problem where updates along most directions provide no curvature information.
In contrast, CoCo aggregates curvature from many pairwise interactions, leading to a well-conditioned Hessian with non-zero curvature in all directions. 
Additionally, the authors of DR showed in~\cite{yang2022inducing} that when representations are already close to the optimal configuration, DR can converge faster than CE. Under similar conditions, and by a transitive argument, CoCo can also converge faster than CE.

\subsubsection{Class dispersion \label{subsub:class_dispersion}}
The constraint induced by DR, namely $h^\top v_c = 1$, does not force representations of samples from the same class to coincide. 
Indeed, for two samples $h_i,h_j$ belonging to class $c$ we have
$
h_i^\top v_c = 1 
$
 and 
$
h_j^\top v_c = 1,
$
which implies
$
(h_i - h_j)^\top v_c = 0 .
$
This condition only enforces that the difference $h_i - h_j$ is orthogonal to $v_c$. 
Therefore, all representations of class $c$ lie in the affine hyperplane defined by $h^\top v_c = 1$, but they may still vary freely along the $(q-1)$ directions orthogonal to $v_c$. 
As a consequence, the DR objective does not explicitly constrain the intra-class dispersion.

In contrast, CoCo operates through pairwise similarity constraints between samples. 
Samples belonging to the same class are encouraged to maximize their similarity, while samples from different classes are pushed apart. 
As a result, the CoCo objective implicitly promotes the collapse of representations belonging to the same class toward a common direction on the unit hypersphere.

Under a mild smoothness assumption on the embedding function $h$, meaning that samples from the same class produce similar embeddings, these pairwise constraints naturally lead to tighter clustering of same-class representations compared to DR.

\section{Experiments}
\label{sec:experiments}
All the materials and implementations required for replication are available at  \url{https://github.com/BlancaCC/the-wonderful-coco}.

\subsection{Setup}
To evaluate the performance of the proposed framework, we conducted extensive experiments on the standardized OpenML-CC18 benchmarking suite \cite{oml-benchmarking-suites}. This suite contains 72 curated binary and multiclass classification datasets designed to provide a diverse and reliable evaluation environment for machine learning methods.

All datasets satisfy several quality criteria: each dataset contains at least 500 observations, has no perfectly predictive features, and presents a minority-to-majority class ratio larger than $5\%$. These constraints ensure that the tasks are non-trivial and suitable for meaningful benchmarking.

The datasets exhibit substantial diversity in terms of size, dimensionality, and number of classes. The number of samples ranges approximately from $500$ to more than $100\,000$ instances, while the number of features varies from fewer than $10$ to several hundred dimensions depending on the dataset.
The classification tasks include both binary and multiclass problems, with the number of classes typically ranging from $2$ to around $50$.

Model performance is evaluated using stratified cross-validation in order to preserve the class distribution across splits. 
Specifically, we employ a stratified $K$-fold scheme with $K=10$. 
At each iteration, one fold is used as the test set while the remaining folds form the training set. 
This procedure produces $10$ independent train-test evaluations per dataset.

Based on the theory exposed in Sections~\ref{sec:introducing_coco} and~\ref{subsec:art_of_state}  we consider the following models, each employing different techniques and training losses.
Recall that $d$ denotes the dimensionality of each dataset sample, $\numClasses$ is the number of target classes and $q \in \mathbb{N}$ is the dimension output which will be hyperametrized. 
The considered models are:
\begin{itemize}
    \item \textbf{CoCo:} This model employs a neural network $h: \mathbb{R}^d \to \mathbb{R}^q$ trained using CoCo loss ~\eqref{eq:coco-loss} if the ratio of the number of elements of the biggest class divided by the minority class is smaller than 3, otherwise we use balanced CoCo~\eqref{eq:balanced-coco-loss}. 
    
    \textit{Prediction:} Once $h$ is trained, classification is performed using the GMM explained in Section~\ref{sub:prediction_with_coco}.

    \item \textbf{DR:} Following \cite{yang2022inducing}, this network $h: \mathbb{R}^d \to \mathbb{R}^q$ is trained to map samples toward fixed ETF prototypes $v_c$ by minimizing $\mathcal{L}_{\mathrm{DR}}$. Like CoCo, it is a classifier-free approach during training.
    
    \textit{Prediction:} Classification is performed by assigning the sample to the class of the fixed prototype with the highest angular similarity:
    $
    \hat{y} = \arg\max_{c} h(x)^\top v_c.
    $

    \item \textbf{CE:} This is the standard probabilistic baseline. A neural network $f: \mathbb{R}^d \to \mathbb{R}^{\numClasses}$ is trained, where an explicit output layer (a learnable affine transformation) maps the penultimate features to $\numClasses$ logits.
    
    \textit{Prediction:} The model outputs a probability distribution via the softmax function, and the class with the highest logit (posterior probability) is selected: $\hat{y} = \arg\max_{c} f(x)_c.$
    \item \textbf{KSVM:} A Gaussian Kernel Support Vector Machine that solves the dual optimization problem.
    
    \textit{Prediction:} For multiclass problems, we adopt a One-Vs-Rest scheme. The prediction is determined by the classifier that yields the maximum margin (distance to the hyperplane) for the transformed input in the RKHS:
    $$
    \hat{y} = \arg\max_{c} \sum_{i=1}^N \alpha_i^{(c)} y_i^{(c)} K(x_i, x) + b^{(c)}.
    $$
    
    \item \textbf{Random Forest (\cite{hastie2009elements}):} A bagging-based ensemble of decision trees trained on bootstrap samples with random feature selection at each split. Random Forests provide a strong and robust baseline for tabular classification tasks.
    
    \textit{Prediction:} Each tree produces a class prediction and the final label is obtained by majority voting across the ensemble:
    $
    \hat{y} = \arg\max_{c} \sum_{t=1}^{T} \mathbf{1}[f_t(x)=c].
    $

\textit{Prediction:} The model outputs class probabilities obtained from the ensemble of boosted trees, and the predicted label is
$
\hat{y} = \arg\max_{c} p(c \mid x).
$
\end{itemize}

%\paragraph{Hyperparameter optimization}

Hyperparameters are selected using Bayesian optimization implemented with Optuna~\cite{optuna_2019}. 
The objective function performs an inner stratified cross-validation on the training data. 
In particular, the training set of each outer fold is further split using $3$ stratified folds. 
For each hyperparameter configuration proposed by the optimizer, the model is trained on two folds and validated on the remaining fold, repeating this process across the three splits. 
The validation score used by the optimizer is the mean value across these folds.
To avoid performing hyperparameter optimization in each of the ten outer folds for every dataset and model, the search was conducted only in the first outer fold, and the selected hyperparameters were fixed for the remaining folds. As the outer folds share approximately 90\% of the data, the optimal hyperparameters are expected to be similar across them. This approach substantially reduces the computational cost.
\paragraph{Hyperparameter search space}
The search space is:
\begin{enumerate}[(a)]
\item \textbf{SVM:} $C \in [10^{-3},10^{5}]$ (log scale).
\item \textbf{Kernel:} Gaussian (RBF).
\item \textbf{Kernel width:} $\gamma \in [4^{-4}/d,\,4^{6}/d]$ (log scale), where $d$ is the input dimension.
\item \textbf{Hidden layers:} $\upsilon \in [1,4]$.
\item \textbf{Hidden units:} $2^{\ell}$ with $\ell \in [\ell_{\min},\ell_{\max}]$, where $\ell_{\min}=\max(\min(\lfloor\log_2(d/2)\rfloor-1,1),2)$ and $\ell_{\max}=\max(\min(\lfloor\log_2(2d)\rfloor+1,5),10)$.
\item \textbf{Batch size:} $m \in [2^4,\,2^{\min(\lfloor\log_2(n/2)\rfloor,9)}]$, where $n$ is the number of training samples.
\item \textbf{Learning rate:} $\eta \in [10^{-5},10^{-2}]$ (log scale).
\item \textbf{Weight decay:} $\lambda \in [10^{-5},10^{-1}]$ (log scale).
\item \textbf{Dropout:} First it decided if it is used, before which layer and the rate $p\in[0.01,0.6]$ if used.
% \item \textbf{CoCo parameter:} $\alpha \in [0.5,10]$.
\item Random Forest \textbf{number of features} in $[1, d]$ where $d$ is the input dimension.
\item Random Forest \textbf{number of estimators} is $500$. 
\end{enumerate}
For the neural models, training was performed for 500 epochs without early stopping. During hyperparameter optimization, each model was allocated $10 \times t$ trials, where $t$ is the number of hyperparameters to be tuned.
Models use the following subsets of the search space: KSVM (a–c), CoCo (d–i), CE (d–i), DR (d–i) and Random Forest (j-k). 
Note that the search space used for Random Forest is the same as the one used in~\cite{fernandez2014we}, in which it ranked first across the studied models.

\subsubsection{Metrics~\label{sec:metric}}
 
The performance of each loss is evaluated by the following metrics: 
\begin{itemize}
    \item \textbf{Balanced accuracy (BA):} For a classification problem with $\numClasses$ classes, let $\mathrm{TP}_c$ and $\mathrm{FN}_c$ denote the number of true positives and false negatives for class $c$, respectively. The recall of class $c$ is defined as
    $$\mathrm{Recall}_c = \frac{\mathrm{TP}_c}{\mathrm{TP}_c + \mathrm{FN}_c}.$$
The balanced accuracy is then defined as the average of the recalls across all classes:
    $$\mathrm{BalancedAccuracy} := \frac{1}{\numClasses}\sum_{c=1}^{\numClasses} \mathrm{Recall}_c .$$

    \item \textbf{Measuring contracting and collapsing (\textit{dispersity}):} 
In order to quantify class separability together with intra-class compactness, we introduce a metric that we refer to as \emph{dispersity} and follows the formulation proposed in~\cite{papyan2020prevalence}:% and without normalising by the number of classes $C$, it is defined as
\begin{equation*}
    \mathrm{dispersity} := %\frac{1}{\numClasses}
    %\operatorname{Tr}\left(\Sigma_{W}\, \Sigma_{B}^{\dagger}\right),
  \|\Sigma_{W}\|/\|\Sigma_{B}\|
\end{equation*}
where $\Sigma_{W}$ is the within-class covariance and $\Sigma_{B}$ is the between-class
covariance.
%, and $\Sigma_{B}^{\dagger}$ denotes the Moore--Penrose pseudoinverse of $\Sigma_{B}$.  
Given feature vectors $\{h_{i,c}\}$, where $h_{i,c}$ denotes the learned
representation of sample $i$ in class $c$ (specifically, the output of $h$ in the
CoCo formulation, the last hidden-layer activation in the CE
setting, and the network output in DR), and assuming $N_{c}$
samples per class $c$,
let
\begin{equation*}
    \mu_{c}=\frac{1}{N_{c}}\sum_{i=1}^{N_{c}} h_{i,c},
    \qquad
    \mu_{G}=\frac{1}{N}\sum_{c}\sum_{i=1}^{N_{c}} h_{i,c},
\end{equation*}
where $N=\sum_{c=1}^\numClasses N_{c}$ is the total number of samples.
The covariance matrices are defined as:
\begin{align*}
    \Sigma_{T}
    &= \frac{1}{N}\sum_{c}^\numClasses\sum_{i=1}^{N_{c}}
        (h_{i,c} - \mu_{G})(h_{i,c} - \mu_{G})^{\top}, \\[2mm]
    \Sigma_{B}
    &= \frac{1}{C}\sum_{c}^\numClasses
        (\mu_{c} - \mu_{G})(\mu_{c} - \mu_{G})^{\top}, \\[2mm]
    \Sigma_{W}
    &= \frac{1}{N}\sum_{c}^\numClasses\sum_{i=1}^{N_{c}}
        (h_{i,c} - \mu_{c})(h_{i,c} - \mu_{c})^{\top},
\end{align*}
where $C$ is the number of classes.
The quantity $\operatorname{Tr}(\Sigma_{W}\Sigma_{B}^{\dagger})$ measures the
total within-class covariance %after whitening
with respect to the between-class
covariance.  This construction is closely related to the Fisher Linear Discriminant
Analysis criterion~\citep{diaz2019deep, hastie2009elements}, where the trace expression
$\operatorname{Tr}(\Sigma_{T}^{-1}\Sigma_{B})$ quantifies class separability.  
In our case, replacing $\Sigma_{T}^{-1}$ with 
$\Sigma_{B}$ 
%$\Sigma_{B}^{\dagger}$ yields a collapse
metric that decreases as the representation approaches neural collapse: within-class
covariance contracts while class means become increasingly separated.
\end{itemize}

\subsection{Statistical evaluation}
\label{sec:statistical_tests}

All reported results are accompanied by statistical comparisons in order to assess whether the observed performance differences between models are significant. 
Following the recommended methodology for comparing multiple classifiers across multiple datasets proposed by \cite{Demsar2006}, we adopt a two-stage statistical evaluation procedure.

\paragraph{Dataset-wise ranking}

For each dataset, we first compare the models using the results (BA or dispersity) obtained from the $10$ outer folds of the cross-validation procedure. 
For every pair of models $(\mathcal{V}_i,\mathcal{V}_j)$ we perform a paired Wilcoxon signed-rank test~\cite{Wilcoxon1945} on the fold-wise performance differences.
Since multiple pairwise comparisons are performed, the resulting $p$-values are adjusted using the Holm--Bonferroni correction~\cite{Holm1979} to control the family-wise error rate at significance level $\alpha = 0.05$.

Model $\mathcal{V}_i$ is considered significantly better than model $\mathcal{V}_j$ if it achieves a higher mean performance and the corresponding Holm-adjusted $p$-value is smaller than $\alpha$.

Based on these pairwise comparisons, a dataset-specific ranking is constructed using a competition ranking scheme: the rank of a model is defined as $1$ plus the number of models that are significantly better than it. 
When no statistically significant difference is detected between two models, they are assigned the same rank, resulting in possible ties.

\paragraph{Average ranking across datasets}

Once the ranking for each dataset is obtained, we compute the average rank of every model across all datasets. 
These average ranks summarize the global relative performance of the models over the entire benchmark.

To determine whether the observed differences between average ranks are statistically significant, we compute the critical distance using the Nemenyi post-hoc test, again following the framework proposed by Friedman in~\cite{Demsar2006}. 
Two models are considered significantly different if the difference between their average ranks exceeds the Nemenyi critical distance.

This procedure allows us to identify whether one model consistently outperforms the others across datasets while accounting for both dataset variability and multiple hypothesis testing.

\subsection{Results}

\subsubsection{Balanced Accuracy comparison}
% \begin{figure}[t]
% \centering

% \begin{minipage}{0.4\linewidth}
% \centering
% \captionof{table}{Average ranks ($\pm$ std) using BA. Lower is better.}
% \label{tab:ba_ranks}
% \small
% \begin{tabular}{lc}
% \hline
% Model & Rank $\pm$ Std \\
% \hline
% \textbf{CoCo} & $\mathbf{2.12 \pm 0.88}$ \\
% \textbf{KSVM} & $2.59 \pm 1.30$ \\
% \textbf{DR }& $2.62 \pm 1.11$ \\
% RF & $3.18 \pm 1.48$ \\
% CE & $4.49 \pm 0.92$ \\
% \hline
% \end{tabular}
% \end{minipage}
% \hfill
% \begin{minipage}{0.59\linewidth}
% \centering
% \begin{tikzpicture}[scale=0.75]

% % axis
% \draw (0,0) -- (8,0);

% % ticks
% \foreach \x/\lab in {0/1,2/2,4/3,6/4,8/5}
% {
% \draw (\x,0.1) -- (\x,-0.1);
% \node at (\x,-0.4) {\lab};
% }

% \node at (4,-0.9) {\footnotesize Avg Rank};

% % scaled positions
% \def\coco{2.24}
% \def\svm{3.17}
% \def\dr{3.24}
% \def\rf{4.36}
% \def\ce{6.98}

% % connectors
% \draw (\coco,0) -- (\coco,0.9);
% \draw (\svm,0) -- (\svm,1.3);
% \draw (\dr,0) -- (\dr,1.7);
% \draw (\rf,0) -- (\rf,2.1);
% \draw (\ce,0) -- (\ce,2.5);

% % labels
% \node[above] at (\coco,0.9) {\scriptsize CoCo};
% \node[above] at (\svm,1.3) {\scriptsize KSVM};
% \node[above] at (\dr,1.7) {\scriptsize DR};
% \node[above] at (\rf,2.1) {\scriptsize RF};
% \node[above] at (\ce,2.5) {\scriptsize CE};

% % groups
% \draw[thick] (\coco,0.6) -- (\dr,0.6);
% \draw[thick] (\svm,0.8) -- (\rf,0.8);

% % CD bar
% \draw[thick] (0,3.0) -- (1.6,3.0);
% \node at (0.8,3.3) {\scriptsize CD = 0.72};

% \end{tikzpicture}

% \caption{Critical difference diagram (Nemenyi, $\alpha=0.05$).}
% \label{fig:cd}
% \end{minipage}

% \end{figure}

\begin{figure}[t]
\centering
% --- LADO IZQUIERDO: LA TABLA (Subfigura a) ---
\begin{subfigure}[b]{0.40\linewidth}
    \centering
    \small % Tamaño de fuente adecuado para tablas en dos columnas
    \begin{tabular}{lc}
    \hline
    Model & Rank $\pm$ Std \\
    \hline
    \textbf{CoCo} & $\mathbf{1.97 \pm 0.97}$ \\
    \textbf{DR} & $\mathbf{2.65 \pm 1.05}$ \\
    \textbf{KSVM} & $\mathbf{2.65 \pm 1.24}$ \\
    RF & $3.20 \pm 1.41$ \\
    CE & $4.53 \pm 0.87$ \\
    \hline
    \end{tabular}
    %\vspace{1.2cm} % Ajuste manual para alinear la subcaption con la de la derecha
    \caption{Average ranks ($\pm$ std).}
    \label{tab:ba_ranks}
\end{subfigure}
\hfill
% --- LADO DERECHO: EL DIAGRAMA (Subfigura b) ---
\begin{subfigure}[b]{0.58\linewidth}
    \centering
    \begin{tikzpicture}[scale=0.72] % Reducido ligeramente para evitar desbordamiento

    % axis
    \draw (0,0) -- (8,0);

    % ticks
    \foreach \x/\lab in {0/1,2/2,4/3,6/4,8/5}
    {
    \draw (\x,0.1) -- (\x,-0.1);
    \node at (\x,-0.4) {\lab};
    }

    \node at (4,-0.9) {\footnotesize Avg Rank};

    % scaled positions (x = 2*(rank-1))
    \def\coco{1.94}
    \def\dr{3.30}
    \def\svm{3.30}
    \def\rf{4.40}
    \def\ce{7.06}

    % connectors
    \draw (\coco,0) -- (\coco,0.9);
    \draw (\dr,0) -- (\dr,1.3);
    \draw (\svm,0) -- (\svm,1.7);
    \draw (\rf,0) -- (\rf,2.1);
    \draw (\ce,0) -- (\ce,2.5);

    % labels
    \node[above] at (\coco,0.9) {\scriptsize CoCo};
    \node[above] at (\dr,1.3) {\scriptsize DR};
    \node[above] at (\svm,1.7) {\scriptsize KSVM};
    \node[above] at (\rf,2.1) {\scriptsize RF};
    \node[above] at (\ce,2.5) {\scriptsize CE};

    % groups
    \draw[thick] (\coco,0.6) -- (\svm,0.6);
    \draw[thick] (\dr,0.8) -- (\rf,0.8);

    % CD bar
    \draw[thick] (0,3.0) -- (1.44,3.0);
    \node at (0.72,3.3) {\scriptsize CD = 0.72};

    \end{tikzpicture}
    \caption{CD diagram (Nemenyi test).}
    \label{fig:cd_dispersity}
\end{subfigure}

\caption{Statistical comparison using BA: (a) average ranks (lower is better) and (b) critical difference diagram at $\alpha=0.05$.}
\label{fig:full_ba_analysis}
\end{figure}
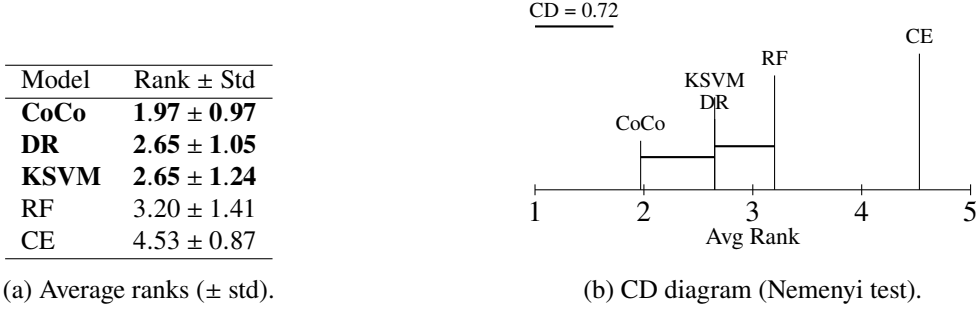

Table~\ref{tab:ba_ranks} reports the average ranks obtained by each model across datasets when evaluated using BA.
The statistical comparison follows the methodology described in Section~\ref{sec:statistical_tests}.
The Friedman test indicates statistically significant differences between the models ($p=2.17\times10^{-19}$).

% \begin{figure}[t]
% \centering
% \begin{minipage}{0.4\linewidth}
% \centering
% \captionof{table}{Average ranks ($\pm$ std) using BA. Lower is better.}
% \label{tab:ba_ranks}
% \small
% \begin{tabular}{lc}
% \hline
% Model & Rank $\pm$ Std \\
% \hline
% \textbf{CoCo} & $\mathbf{1.97 \pm 0.97}$ \\
% \textbf{DR} & $2.65 \pm 1.05$ \\
% \textbf{KSVM} & $2.65 \pm 1.24$ \\
% RF & $3.20 \pm 1.41$ \\
% CE & $4.53 \pm 0.87$ \\
% \hline
% \end{tabular}
% \end{minipage}
% \hfill
% \begin{minipage}{0.59\linewidth}
% \centering
% \begin{tikzpicture}[scale=0.75]

% % axis
% \draw (0,0) -- (8,0);

% % ticks
% \foreach \x/\lab in {0/1,2/2,4/3,6/4,8/5}
% {
% \draw (\x,0.1) -- (\x,-0.1);
% \node at (\x,-0.4) {\lab};
% }

% \node at (4,-0.9) {\footnotesize Avg Rank};

% % scaled positions (x = 2*(rank-1))
% \def\coco{1.94}
% \def\dr{3.30}
% \def\svm{3.30}
% \def\rf{4.40}
% \def\ce{7.06}

% % connectors
% \draw (\coco,0) -- (\coco,0.9);
% \draw (\dr,0) -- (\dr,1.3);
% \draw (\svm,0) -- (\svm,1.7);
% \draw (\rf,0) -- (\rf,2.1);
% \draw (\ce,0) -- (\ce,2.5);

% % labels
% \node[above] at (\coco,0.9) {\scriptsize CoCo};
% \node[above] at (\dr,1.3) {\scriptsize DR};
% \node[above] at (\svm,1.7) {\scriptsize KSVM};
% \node[above] at (\rf,2.1) {\scriptsize RF};
% \node[above] at (\ce,2.5) {\scriptsize CE};

% % groups
% \draw[thick] (\coco,0.6) -- (\svm,0.6);
% \draw[thick] (\dr,0.8) -- (\rf,0.8);

% % CD bar (scaled: 2 * 0.72 = 1.44)
% \draw[thick] (0,3.0) -- (1.44,3.0);
% \node at (0.72,3.3) {\scriptsize CD = 0.72};

% \end{tikzpicture}

% \caption{Critical difference diagram (Nemenyi, $\alpha=0.05$).}
% \label{fig:cd_dispersity}
% \end{minipage}
% \end{figure}

CoCo achieves the best overall performance with an average rank of $1.97 \pm 0.97$, followed by DR and KSVM (both $2.65 \pm 1.05$ and $2.65 \pm 1.24$, respectively). 
Random Forest (RF) obtains an intermediate position ($3.20 \pm 1.41$), while the neural network trained with cross-entropy shows the worst performance ($4.53 \pm 0.87$).
Using the Nemenyi post-hoc test, the critical distance is $CD = 0.72$, Figure~\ref{fig:cd_dispersity}.
%The results are summarized in Figure~\ref{fig:cd}.
According to this threshold, CoCo significantly outperforms Random Forest and CE. 
In contrast, the differences between CoCo, Dot Regression, and KSVM are not statistically significant, although they are close to the significance threshold.

Additionally, CE is significantly worse than all other models, indicating that standard neural network training with cross-entropy is less effective than the alternative approaches considered in this study for tabular classification.

These results are consistent with previous empirical studies on tabular learning~\cite{fernandez2014we}, which also suggest that the leading models typically achieve comparable performance levels across datasets.
In this context, our findings also serve as a sanity check, confirming that CoCo performance is comparable to well-established methods.

% end balanced accuracy
\subsubsection{Dispersity: collapsing and contrasting measure}

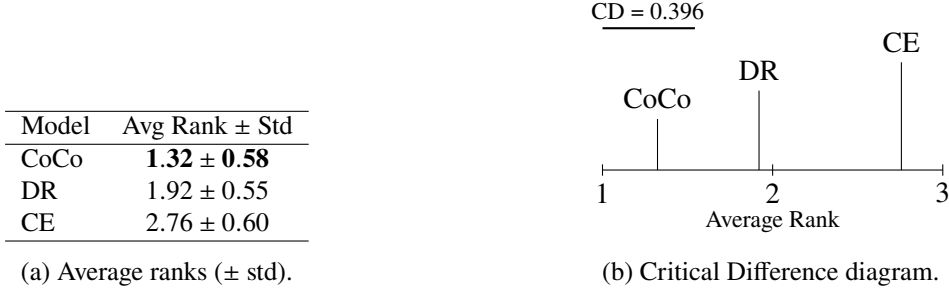
\begin{figure}[t]
\centering
% --- Subfigura A: La Tabla ---
\begin{subfigure}[b]{0.42\linewidth}
    \centering
    \small
    \begin{tabular}{lc}
    \hline
    Model & Avg Rank $\pm$ Std \\
    \hline
    CoCo & $\mathbf{1.32 \pm 0.58}$ \\
    DR & $1.92 \pm 0.55$ \\
    CE & $2.76 \pm 0.60$ \\
    \hline
    \end{tabular}
    \caption{Average ranks ($\pm$ std).}
    \label{tab:dispersion_sub}
\end{subfigure}
\hfill
% --- Subfigura B: El Diagrama ---
\begin{subfigure}[b]{0.56\linewidth}
    \centering
    \begin{tikzpicture}[scale=0.75] % Un poco más pequeño para que respire
        % axis
\draw (0,0) -- (6,0);

% ticks
\foreach \x/\lab in {0/1,3/2,6/3}
{
\draw (\x,0.1) -- (\x,-0.1);
\node at (\x,-0.4) {\lab};
}

\node at (3,-0.9) {\footnotesize Average Rank};

% positions
\def\coco{0.97}
\def\dr{2.76}
\def\ce{5.27}

\draw (\coco,0) -- (\coco,0.9);
\draw (\dr,0) -- (\dr,1.4);
\draw (\ce,0) -- (\ce,1.9);

\node[above] at (\coco,0.9) {CoCo};
\node[above] at (\dr,1.4) {DR};
\node[above] at (\ce,1.9) {CE};

% CD
\draw[thick] (0,2.5) -- (1.63,2.5);
\node at (0.8,2.8) {\footnotesize CD = 0.396};
    \end{tikzpicture}
    \caption{Critical Difference diagram.}
    \label{fig:dispersion_cd_sub}
\end{subfigure}

\caption{Dispersity metric results: (a) average ranks and (b) Nemenyi test CD diagram ($\alpha=0.05$).}
\label{fig:combined_dispersion}
\end{figure}

To evaluate the dispersion of the learned representations, we measure the ratio between intra-class variance and inter-class separation (\textit{dispersity}) defined in Section~\ref{sec:metric}.
Note that the evaluation can only be performed for models that produce embeddings. In our case, these correspond to the representations from the CE last hidden layer and the outputs of DR and CoCo. The results are shown in Figure~\ref{fig:combined_dispersion}.%Table~\ref{tab:dispersion}.
The statistical comparison follows the methodology described in Section~\ref{sec:statistical_tests}.% and it is shown in Figure~\ref{fig:combined_dispersion}.
The Friedman test indicates statistically significant differences between the models ($p=4.76\times10^{-9}$).

% \begin{figure}[t]
% \centering
% \begin{minipage}{0.40\linewidth}
% \centering
% \captionof{table}{Average ranks ($\pm$ std) for the \textit{dispersity metric}. Lower ranks indicate better intra class collapse and inter class separation.}
% \label{tab:dispersion}
% \begin{tabular}{lc}
% \hline
% Model & Avg Rank $\pm$ Std \\
% \hline
% CoCo & $\mathbf{1.32 \pm 0.58}$ \\
% DR & $1.92 \pm 0.55$ \\
% CE & $2.76 \pm 0.60$ \\
% \hline
% \end{tabular}
% \end{minipage}
% \hfill
% \begin{minipage}{0.59\linewidth}
% \centering

% \begin{tikzpicture}[scale=0.85]

% % axis
% \draw (0,0) -- (6,0);

% % ticks
% \foreach \x/\lab in {0/1,3/2,6/3}
% {
% \draw (\x,0.1) -- (\x,-0.1);
% \node at (\x,-0.4) {\lab};
% }

% \node at (3,-0.9) {\footnotesize Average Rank};

% % positions
% \def\coco{0.97}
% \def\dr{2.76}
% \def\ce{5.27}

% \draw (\coco,0) -- (\coco,0.9);
% \draw (\dr,0) -- (\dr,1.4);
% \draw (\ce,0) -- (\ce,1.9);

% \node[above] at (\coco,0.9) {CoCo};
% \node[above] at (\dr,1.4) {DR};
% \node[above] at (\ce,1.9) {CE};

% % CD
% \draw[thick] (0,2.5) -- (1.63,2.5);
% \node at (0.8,2.8) {\footnotesize CD = 0.396};

% \end{tikzpicture}

% \caption{Critical Difference diagram for the dispersity metric (Nemenyi test, $\alpha=0.05$).}
% \label{fig:dispersion_cd}

% \end{minipage}

% \end{figure}
The Nemenyi post-hoc test ($CD=0.396$) shows that CoCo significantly outperforms both DR and CE, while DR also significantly outperforms CE. 
These results indicate that CoCo produces the most compact class representations among the evaluated methods.

This empirical observation is consistent with the theoretical analysis presented in Section~\ref{subsub:class_dispersion}, where we showed that the CoCo objective explicitly promotes the collapse of same-class representations while maintaining separation between different classes. 
Dot Regression also improves clustering compared to CE, although the effect is weaker since DR only constrains representations to lie on class-specific hyperplanes, allowing variability along directions orthogonal to the prototype.

Looking at this ranking, the relatively poorer clustering behaviour observed for CE may partly reflect the fact that this training regime has not yet exhibited neural collapse in its last hidden layer. This behavior could be due to the nature of the datasets or to an insufficient number of training epochs  (500). 

\subsubsection{Early convergence}

The work of~\cite{yang2022inducing} shows that when representations are already close to the optimal configuration, DR can converge faster than CE. However, evaluating performance only near the optimum does not reflect the usual training scenario, even when pretrained models are used. For this reason, we analyze convergence behaviour from random initialization and evaluate how quickly each method reaches competitive performance in terms of BA.

In this experiment, all methods share the same architecture and training configuration to ensure a fair comparison.
Hyperparameters are chosen to provide stable training across methods without favoring any particular loss.
For CoCo and DR we use a two-layer neural network defined as
$h(x) = \tanh\!\left(W_2 \,\mathrm{ReLU}(W_1 x + w_1) + w_2\right),$
where
$
W_1 \in \R^{u_1 \times d}, \quad
W_2 \in \R^{u_2 \times u_1}, \quad
w_1 \in \R^{u_1}, \quad
w_2 \in \R^{u_2}.
$

The hidden dimensions $u_1$ and $u_2$ depend on the input dimension $d$ and the number of classes $C$:
\[
u_1 = \min(\max(10d, 4C), 1024), \qquad
u_2 = \max(\min(3d, 2C), 8).
\]

For CE, we add a final linear classifier, that is, 
$f(x) = \mathrm{softmax}(Wh(x)),$
where $W \in \R^{C \times u_2}$.
% The model is for CoCo and DR a two-layer neural network, $h(x) = \tanh(W_2 \text{relu}(W_1 x + w_1)+w_2)$, whose dimensions $W_1 \in R^{u_1 \times d}$  
% $W_2 \in R^{u_2 \times u_1}$
% where $u_1, u_2$ depend on the input dimension $d$ and the number of classes $C$:
% \begin{equation*}
% u_1 = \min(\max(10d, 4C), 1024), \qquad
% u_2 = \max(\min(3d, 2C), 8).
% \end{equation*}
% For CE we add the final linear classifier, this is, $f(x) = \text{softmax}(W h(x))$ where $W\in R^{C \times u_2}$.
The remaining parameters are fixed across experiments: learning rate $10^{-4}$, batch size $8$
%, CoCo parameter $\alpha = 1$, 
and no regularization. A small batch size is used to allow a finer observation of the early training dynamics.

An important aspect of this comparison is that the losses differ in the number of terms contributing to the gradients during each minibatch update.
For a minibatch of size $m$, CE and DR process $m$ independent samples, producing $m$ terms in the loss function that contribute to the gradient update.
In contrast, CoCo operates on pairwise relations between samples, generating $m^2$ interaction terms.

To ensure a fair comparison, convergence is measured with respect to the total number of gradient terms rather than the number of iterations.
After each minibatch update, BA in test is evaluated and the cumulative number of terms is updated as
\begin{equation*}
N_{\text{terms}} \leftarrow N_{\text{terms}} + m
\quad \text{for CE and DR},
\end{equation*}
\begin{equation*}
N_{\text{terms}} \leftarrow N_{\text{terms}} + m^2
\quad \text{for CoCo}.
\end{equation*}
This protocol ensures that all methods are compared under the same amount of gradient information.

As a result, Table~\ref{tab:early_convergence} and  Figure~\ref{fig:early_cd} report the average significant ranking of each model across datasets as a function of $n_{\text{terms}}$, together with the corresponding critical distance obtained from the Nemenyi test.
\begin{table}[t]
\centering
\caption{Average ranks ($\pm$ std) for early convergence analysis measured with respect to the number of gradient terms $n_{\text{terms}}$. Lower ranks indicate better performance.}
\label{tab:early_convergence}
\begin{tabular}{lccccc}
\hline
Model / $n_{\text{terms}}$ & 64 & 128 & 256 & 512 & 1024 \\
\hline
CoCo & $\mathbf{1.0 \pm 0.0}$ & $\mathbf{1.0 \pm 0.0}$ & $\mathbf{1.0 \pm 0.1}$ & $\mathbf{1.2 \pm 0.4}$  & $\mathbf{1.3 \pm 0.6}$\\
DR & $2.4 \pm 0.5$ & $2.3 \pm 0.5$ & $2.2 \pm 0.5$ & $2.1 \pm 0.7$ & $1.9 \pm 0.6$ \\
CE & $2.6 \pm 0.5$ & $2.7 \pm 0.5$ & $2.8 \pm 0.4$ & $2.7 \pm 0.5$ & $2.7 \pm 0.5$\\
\hline
CD ($\alpha=0.05$) & $0.4$ & $0.4$ & $0.4$ & $0.4$ & $0.4$ \\
\hline
\end{tabular}
\end{table}
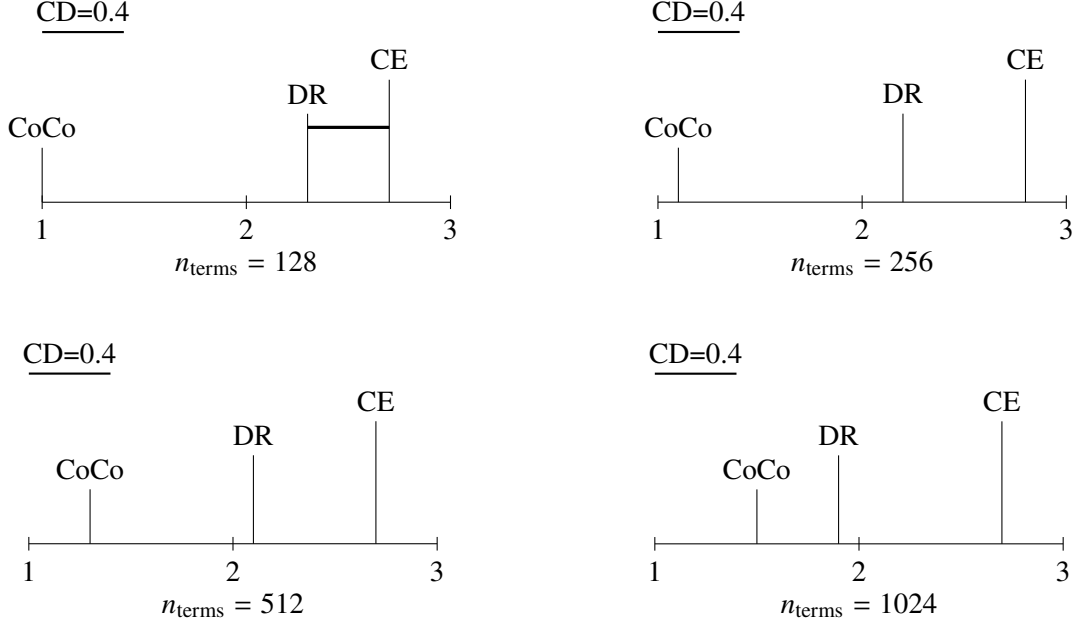
\begin{figure}[t]
\centering

% -------- ROW 1 --------
\begin{minipage}{0.48\linewidth}
\centering
\begin{tikzpicture}[scale=0.9]

\draw (0,0)--(6,0);
\foreach \x/\lab in {0/1,3/2,6/3}
{
\draw (\x,0.1)--(\x,-0.1);
\node at (\x,-0.4){\lab};
}

\node at (3,-0.9){$n_{\text{terms}}=128$};

\def\coco{0}
\def\dr{3.9}
\def\ce{5.1}

\draw (\coco,0)--(\coco,0.8);
\draw (\dr,0)--(\dr,1.3);
\draw (\ce,0)--(\ce,1.8);

\node[above] at (\coco,0.8){CoCo};
\node[above] at (\dr,1.3){DR};
\node[above] at (\ce,1.8){CE};

% non-significant group
\draw[very thick] (\dr,1.1)--(\ce,1.1);

\draw[thick] (0,2.5)--(1.2,2.5);
\node at (0.6,2.8){CD=0.4};

\end{tikzpicture}
\end{minipage}
\hfill
\begin{minipage}{0.48\linewidth}
\centering
\begin{tikzpicture}[scale=0.9]

\draw (0,0)--(6,0);
\foreach \x/\lab in {0/1,3/2,6/3}
{
\draw (\x,0.1)--(\x,-0.1);
\node at (\x,-0.4){\lab};
}

\node at (3,-0.9){$n_{\text{terms}}=256$};

\def\coco{0.3}
\def\dr{3.6}
\def\ce{5.4}

\draw (\coco,0)--(\coco,0.8);
\draw (\dr,0)--(\dr,1.3);
\draw (\ce,0)--(\ce,1.8);

\node[above] at (\coco,0.8){CoCo};
\node[above] at (\dr,1.3){DR};
\node[above] at (\ce,1.8){CE};

\draw[thick] (0,2.5)--(1.2,2.5);
\node at (0.6,2.8){CD=0.4};

\end{tikzpicture}
\end{minipage}

\vspace{0.6cm}

% -------- ROW 2 --------
\begin{minipage}{0.48\linewidth}
\centering
\begin{tikzpicture}[scale=0.9]

\draw (0,0)--(6,0);
\foreach \x/\lab in {0/1,3/2,6/3}
{
\draw (\x,0.1)--(\x,-0.1);
\node at (\x,-0.4){\lab};
}

\node at (3,-0.9){$n_{\text{terms}}=512$};

\def\coco{0.9}
\def\dr{3.3}
\def\ce{5.1}

\draw (\coco,0)--(\coco,0.8);
\draw (\dr,0)--(\dr,1.3);
\draw (\ce,0)--(\ce,1.8);

\node[above] at (\coco,0.8){CoCo};
\node[above] at (\dr,1.3){DR};
\node[above] at (\ce,1.8){CE};

\draw[thick] (0,2.5)--(1.2,2.5);
\node at (0.6,2.8){CD=0.4};

\end{tikzpicture}
\end{minipage}
\hfill
\begin{minipage}{0.48\linewidth}
\centering
\begin{tikzpicture}[scale=0.9]

\draw (0,0)--(6,0);
\foreach \x/\lab in {0/1,3/2,6/3}
{
\draw (\x,0.1)--(\x,-0.1);
\node at (\x,-0.4){\lab};
}

\node at (3,-0.9){$n_{\text{terms}}=1024$};

\def\coco{1.5}
\def\dr{2.7}
\def\ce{5.1}

\draw (\coco,0)--(\coco,0.8);
\draw (\dr,0)--(\dr,1.3);
\draw (\ce,0)--(\ce,1.8);

\node[above] at (\coco,0.8){CoCo};
\node[above] at (\dr,1.3){DR};
\node[above] at (\ce,1.8){CE};

\draw[thick] (0,2.5)--(1.2,2.5);
\node at (0.6,2.8){CD=0.4};

\end{tikzpicture}
\end{minipage}

\caption{Demšar critical difference diagrams for early convergence measured with respect to the number of gradient terms.}
\label{fig:early_cd}

\end{figure}

% \begin{table}[t]
% \centering
% \caption{Average ranks ($\pm$ std) for early convergence analysis measured with respect to the number of gradient terms $n_{\text{terms}}$. Lower ranks indicate better performance.}
% \label{tab:early_convergence}
% \begin{tabular}{lccccc}
% \hline
% Model / $n_{\text{terms}}$ & 64 & 128 & 256 & 512 & 1024 \\
% \hline
% CoCo & $\mathbf{1.00 \pm 0.00}$ & $\mathbf{1.00 \pm 0.00}$ & $\mathbf{1.01 \pm 0.12}$ & $\mathbf{1.16 \pm 0.41}$  & $\mathbf{1.30 \pm 0.58}$\\
% DR & $2.40 \pm 0.49$ & $2.31 \pm 0.46$ & $2.19 \pm 0.45$ & $2.12 \pm 0.65$ & $1.93 \pm 0.64$ \\
% CE & $2.57 \pm 0.53$ & $2.69 \pm 0.46$ & $2.76 \pm 0.42$ & $2.69 \pm 0.45$ & $2.73 \pm 0.46$\\
% \hline
% CD ($\alpha=0.05$) & $0.396$ & $0.396$ & $0.396$ & $0.396$ & $0.396$ \\
% \hline
% \end{tabular}
% \end{table}
Across all considered values of $n_{\text{terms}}$, CoCo consistently achieves the best average rank. 
In particular, for very small computational budgets ($n_{\text{terms}}=64$ and $128$), CoCo obtains rank $1$ on all datasets, indicating that it provides the best performance during the earliest stages of training.

The statistical analysis confirms that these differences are significant. 
Using the Nemenyi post-hoc test with $\alpha=0.05$, the critical distance is $CD = 0.4$. 
For all tested values of $n_{\text{terms}}$, the difference between CoCo and both DR and CE exceeds this threshold, indicating that CoCo significantly outperforms the other methods during the early training regime.

As the number of terms increases, the gap between the methods gradually decreases. 
In particular, for $n_{\text{terms}}=512$, some datasets start to show comparable performance between CoCo and DR, which explains the slight increase in the average rank of CoCo. 
Nevertheless, CoCo remains the best-ranked model overall.

For $n_{\text{terms}}=1024$, this tendency becomes more pronounced. 
While CoCo still achieves the best average rank, the difference with DR has reduced. In contrast, CE remains stable but significantly worse than both CoCo and DR.

Interestingly, DR consistently improves its average rank as $n_{\text{terms}}$ increases, while CE shows a mild deterioration. This behavior is consistent with the observations reported in~\cite{yang2022inducing}, where DR is shown to accelerate the convergence of CE once the representations approach the optimal configuration.
% For all tested values of $n_{\text{terms}}$, the difference between CoCo and both DR and CE exceeds this threshold, indicating that CoCo significantly outperforms the other methods during the early training regime.

% As the number of terms increases, the gap between the methods gradually decreases. 
% In particular, for $n_{\text{terms}}=512$, some datasets start to show comparable performance between CoCo and DR, which explains the slight increase in the average rank of CoCo. 
% Nevertheless, CoCo remains the best-ranked model overall.
% Interestingly, although DR and CE are not significantly different according to the Nemenyi test, a slight trend can still be observed in their rankings as $n_{\text{terms}}$ increases.
% In particular, DR gradually improves its average rank, while CE shows a mild deterioration. Although these differences are not statistically significant, this behavior is consistent with the observations reported in~\cite{yang2022inducing}, where DR is shown to accelerate the convergence of CE once the representations approach the optimal configuration. 

These results provide empirical evidence supporting the theoretical analysis presented in Section~\ref{sub:positioning_coco}. 
In particular, the combination of closer initialization and more informative gradients allows CoCo to reach competitive representations significantly earlier than DR and CE.

\subsubsection{Improvement of Balanced CoCo over CoCo as a function of dataset imbalance}

\begin{figure}[t]
\centering
\includegraphics[width=0.7\linewidth]{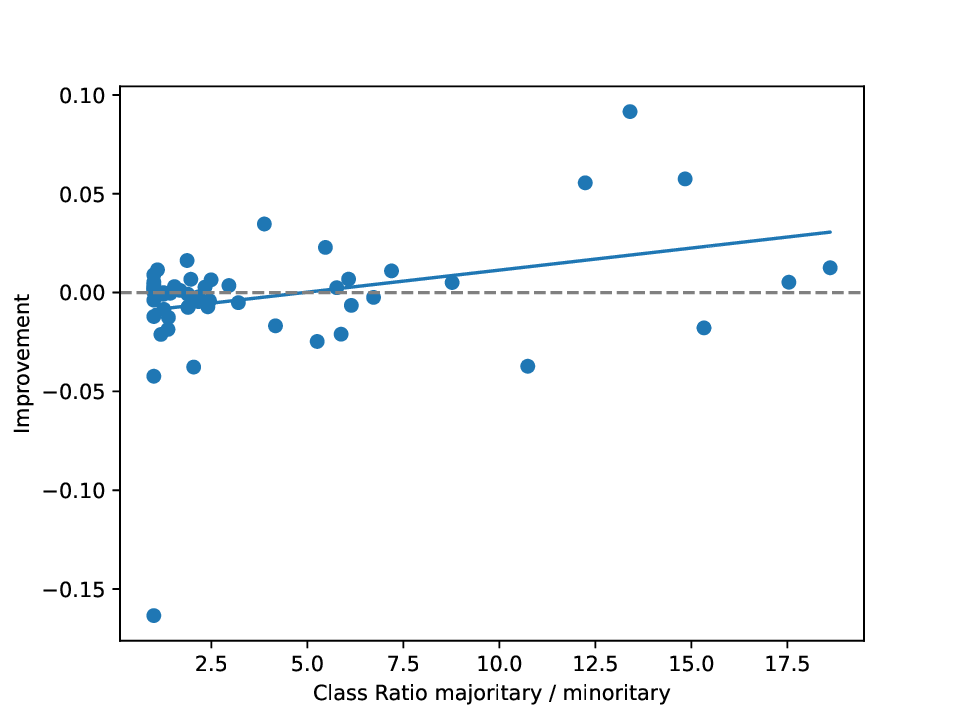}
\caption{Improvement of Balanced CoCo over CoCo as a function of dataset imbalance, measured by the balanced ratio. The solid line represents the linear regression trend.}
\label{fig:imbalance_scatter}
\end{figure}

To further analyze the behavior of CoCo under class imbalance, we conducted an additional experiment comparing the standard CoCo~\eqref{eq:coco-loss} model against its balanced variant~\eqref{eq:balanced-coco-loss}.
For each dataset, we computed the performance in BA difference (improvement) between both models, defined as the difference in average test score across inner folds.
For both loss functions, the hyperparameter optimization procedure described at the beginning of this experimental section was independently applied to ensure a fair comparison.
This improvement was then related to the dataset imbalance, measured through the balanced ratio.

Figure~\ref{fig:imbalance_scatter} illustrates the relationship between improvement and imbalance. Each point corresponds to a dataset, and a linear regression line is included to highlight the overall trend.

The results suggest a weak positive relationship between imbalance and the relative performance of the balanced variant. Specifically, the estimated slope of the linear regression is $0.0022$, indicating that the improvement tends to increase slightly as the imbalance grows. However, this effect is small in magnitude.

This observation is further supported by the Spearman correlation coefficient~\cite{spearman1961proof} ($\rho = 0.2$), which indicates a low positive monotonic association. 
Importantly, the correlation is not statistically significant ($p = 0.14$). However, the distribution of points in Figure~\ref{fig:imbalance_scatter} suggests that when the class ratio is close to 1, the variance of the results is generally low and concentrated around zero.
Moreover, as the class ratio increases, the variability also grows, with a tendency toward more favorable outcomes for the balanced method~\eqref{eq:balanced-coco-loss}. 

Overall, these results indicate that, although the balanced version of CoCo may provide slight benefits in more imbalanced scenarios, the effect is weak and not statistically significant for the datasets considered in this study.
Therefore, imbalance alone does not appear to be a decisive factor in determining the relative performance between both variants in this setting.
A direction for future work would be to focus specifically on moderately to highly imbalanced datasets, where the potential advantages of the balanced formulation may become more pronounced.
\section{Conclusions\label{sec:conclusions}}

In this work, we propose two new loss functions, CoCo and class balanced CoCo, motivated by the limitations of using fixed Equiangular Tight Frames (ETF) as the optimal structure for maximizing angular separation. 
The solutions of CoCo variants are functions that map classification data into a new feature space that satisfies the following properties: 
(i) they are normalized; 
(ii) data from the same class tend to collapse into (or, in practice, cluster around) a single vector; 
(iii) different classes are separated by the maximum distance allowed.
These properties are useful in tasks such as improving classification predictions, building convenient embeddings and even accelerating convergence.

We compared CoCo with Dot Regression (DR), Cross-Entropy (CE), KSVM and Random Forest.
KSVM, Random Forest and CE are the baseline state-of-the-art models for prediction quality, while DR targets embedding quality.
We showed that neural networks trained with CoCo can potentially converge faster to a solution compared to other methods. We also demonstrated that the discriminative quality of their embeddings is superior.

The experimental evaluation spans diverse datasets, focusing on prediction accuracy, representation quality, and convergence speed.

To measure prediction accuracy, we studied the mean balanced accuracy, and in this case, CoCo loss is within the top models without significant differences among then. 
In order to measure the collapse and contrast effect, we used the \textit{dispersity} metric, which is unbounded and non negative, being the best score 0 (the lower, the better).
The best \textit{dispersity} results were obtained with CoCo, closely followed by DR, with CE ranking last. 
These results support our theoretical purpose: 
convergence to the optimal geometrical structure that maximizes collapse and contrasting the data.
This also explains the ranking order.
In CE, the embedding appears as a consequence of the final layer convergence; in DR, the embedding is forced to converge to a fixed ETF;
and in CoCo, we define the constraints that the optimal solution should satisfy while allowing the neural network to find the closest approximation. 

Finally, the convergence analysis highlights the advantage of CoCo losses for rapid discriminability.
CoCo models consistently achieve the best or near-best BA performance across early epochs, demonstrating faster class separation and more robust convergence dynamics than CE and DR.
While CE and DR approaches eventually close the gap with sufficient training, they require substantially more epochs to do so.

In summary, our study demonstrates that CoCo loss provide a robust alternative to traditional objectives by explicitly guiding embeddings toward optimal geometric structures.
It not only achieve superior balanced accuracy on average but also produce more meaningful representations, as evidenced by the \textit{dispersity} analysis.
Overall, CoCo stands out as a robust, theoretically grounded, and empirically validated framework that enhances representation quality, convergence acceleration and keep in top of prediction performance offering a promising direction for future research in loss design and geometric representation learning.

Due to the value of this idea, we have identified the following directions for further work, which we are currently pursuing:
\begin{itemize}
\item Extending the proposed approach to ordinal regression or pure regression problems, for which the target configuration $g$ (defined in Section~\ref{sec:introducing_coco}) needs to be accordingly adapted.
\item Leveraging the clustering property of our loss function to reduce the number of samples, thereby maintaining accuracy while improving computational efficiency in terms of time and memory.
\item Exploiting the model's ability to learn from inherent data structures to apply CoCo in semi-supervised learning problems.
\item Analyze better the effect of CoCo and its balanced version in imbalanced problems and even propouse a mixture model motivate by~\cite{yan2024neural}. 
\end{itemize}

\section*{Acknowledgments}

The authors acknowledge financial support from project
PID2022-139856NB-I00 funded by MCIN/ AEI / 10.13039/501100011033 /
FEDER, UE and project, IDEA-CM (TEC-2024/COM-89) from the Autonomous
Community of Madrid and from the ELLIS Unit Madrid, 
Cátedra UAM-IIC de Ciencia de Datos y Aprendizaje Automático 
and FPI-UAM.
The authors
acknowledge computational support from the Centro de Computación
Científica-Universidad Autónoma de Madrid (CCC-UAM).
The authors thank Estrella Sánchez for valuable discussions and support.

\section*{CRediT authorship contribution statement}
\textbf{Blanca Cano-Camarero:} Conceptualization, Methodology, Software, Validation, Formal analysis, Investigation, Data curation, Writing - original draft, Visualization, Project administration.
\textbf{Ángela Fernández-Pascual:} Conceptualization, Supervision, Writing - review and editing.
\textbf{José R. Dorronsoro:} Conceptualization, Supervision, Writing - review and editing.

\section*{Declaration of Generative AI and AI-assisted technologies in the writing process}
During the preparation of this work the authors used ChatGPT in order to improve the language style. After using this tool/service, the authors reviewed and edited the content as needed and takes full responsibility for the content of the publication

%%%%%%%%%%%%%%%%%%%%%%%%%%%%%%%%%%%%%%%%%%%%%%%%%%%%%%%%%%%%%%%%%%%%%

% --- References ---
% Standard plain or unsrt styles replace the Elsevier-specific style
\bibliographystyle{unsrt} 
\bibliography{bibliography.bib}

% --- Appendix ---
\clearpage
\appendix
\section{CoCo: Optimality and Neural Collapse}
\label{appendix:CoCo_analysis}

\textbf{A. Derivation of correctness and optimality.} 
Suppose we have $\numClasses$ classes with representative unit vectors $v_1, \ldots, v_{\numClasses} \in \R^q$. To maximize pairwise angles, we minimize their common inner product $\gamma \in [-1,1]$. The Gram matrix $G \in \R^{\numClasses \times \numClasses}$ is defined as $G_{ij} = 1$ if $i=j$ and $G_{ij} = \gamma$ otherwise. $G$ is positive semidefinite (PSD) if its eigenvalues are nonnegative \citep{horn2012matrix}.

The matrix $G$ can be written as $G = (1-\gamma)I + \gamma \mathbf{1}\mathbf{1}^\top$. Its eigenvalues are:
\begin{itemize}[leftmargin=*, noitemsep, topsep=0pt]
    \item $\lambda_1 = 1 - \gamma$, with multiplicity $(\numClasses - 1)$.
    \item $\lambda_2 = 1 + (\numClasses - 1)\gamma$, with multiplicity $1$.
\end{itemize}
The PSD condition $\lambda_i \geq 0$ implies $-\frac{1}{\numClasses - 1} \leq \gamma \leq 1$. To maximize angles (minimize $\gamma$), the optimal value is $\gamma^\ast = -\frac{1}{\numClasses - 1}$. Such vectors exist as an embedding of a regular $(\numClasses-1)$-dimensional simplex in $\R^q$ \citep{el2004certain}.

\end{document}